\documentclass{article}

\usepackage{arxiv}
\usepackage[colorlinks=true,linkcolor=blue,citecolor=blue,urlcolor=blue]{hyperref}
\usepackage[utf8]{inputenc} 
\usepackage[T1]{fontenc}
\usepackage[english]{babel}
\usepackage{url}            
\usepackage{booktabs}      
\usepackage{amsfonts}       
\usepackage{nicefrac}       
\usepackage{microtype}   
\usepackage{graphicx}
\usepackage{doi}
\usepackage{caption}
\usepackage{CJKutf8}
\usepackage[most]{tcolorbox}
\usepackage{enumitem}
\usepackage{multirow}
\usepackage{multicol}
\usepackage{threeparttable}
\usepackage{natbib}

\title{CPG-EVAL: A Multi-Tiered Benchmark for Evaluating the Chinese Pedagogical Grammar Competence of Large Language Models}

\author{ \href{https://orcid.org/0009-0001-0410-3694}{\includegraphics[scale=0.06]{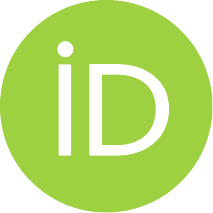}\hspace{1mm}Dong Wang} \\
	Faculty of Education and Integrated Arts and Sciences\\
	Waseda University\\
	Shinjuku-ku, Tokyo, Japan\\
	\texttt{wangdong@aoni.waseda.jp} \\
}

\renewcommand{\shorttitle}{CPG-EVAL}

\begin{document}
\begin{CJK*}{UTF8}{gbsn}
\maketitle

\begin{abstract}
\textbf{Purpose}: The rapid emergence of large language models (LLMs) such as ChatGPT has significantly impacted foreign language education, yet their pedagogical grammar competence remains under-assessed. This paper introduces CPG-EVAL, the first dedicated benchmark specifically designed to evaluate LLMs' knowledge of pedagogical grammar within the context of foreign language instruction.  \textbf{Methodology}: The benchmark comprises five tasks designed to assess grammar recognition, fine-grained grammatical distinction, categorical discrimination, and resistance to linguistic interference.  \textbf{Findings}: Smaller-scale models can succeed in single language instance tasks, but struggle with multiple instance tasks and interference from confusing instances. Larger-scale models show better resistance to interference but still have significant room for accuracy improvement. The evaluation indicates the need for better instructional alignment and more rigorous benchmarks, to effectively guide the deployment of LLMs in educational contexts. \textbf{Value}: This study offers the first specialized, theory-driven, multi-tiered benchmark framework for systematically evaluating LLMs\textquotesingle{} pedagogical grammar competence in Chinese language teaching contexts. CPG-EVAL not only provides empirical insights for educators, policymakers, and model developers to better gauge AI\textquotesingle s current abilities in educational settings, but also lays the groundwork for future research on improving model alignment, enhancing educational suitability, and ensuring informed decision-making concerning LLM integration in foreign language instruction.\footnote{ The CPG-EVAL data and evaluation code are available at https://github.com/wd-github-2017/CPG-EVAL}
\end{abstract}


\section{Introduction}

\textbf{What would be the consequences if educators were allowed to instruct students without undergoing any certification or assessment?}

This situation seems irrational, but is quietly becoming a reality within the field of foreign language education. However, the "educators" in this context are not human teachers but rather large language models (LLMs) such as ChatGPT and DeepSeek, which represent the generative artificial intelligence technologies that are currently so prominent.

In recent years, rapid advancements in generative models have significantly expanded the practical applications of LLMs in foreign language education. Models like ChatGPT have begun to fulfill diverse roles within language learning, including supporting autonomous learning, generating adaptive content, and aiding teachers in instructional design and research processes \citep{karatas2024IncorporatingAIForeign,li2024SystematicReviewFirst}. This trend reflects high expectations surrounding their capabilities for natural language generation and reasoning.

Existing research has started to explore the applicability of LLMs across various language education scenarios, such as writing error correction, conversation practice, and instructional resource development \citep{bin-hady2023ExploringDimensionsChatGPT,kohnke2023ChatGPTLanguageTeaching,liu2024MeasuringEFLLearners}. Historically, benchmarks in natural language processing and artificial intelligence research have served not only as performance measurement tools. They have also functioned as strategic platforms guiding technological development from an expert and sustainability-oriented perspective \citep{srivastava2022beyond,liang2023HolisticEvaluationLanguage,zhuang2023through,chang2024survey}. However, a systematic, education-specific evaluation methodology rooted in the expertise of language education professionals for assessing LLMs has not been developed thus far.

To bridge this gap, this study introduces the first systematic evaluation framework focused on  LLMs’ knowledge of pedagogical grammar, which we call the Chinese Pedagogical Grammar Evaluation (CPG-EVAL). Grounded in a pedagogical grammar classification system that is validated through extensive instructional practice in Teaching Chinese as a Foreign Language, CPG-EVAL provides a comprehensive benchmark for the instructional process. It is designed to evaluate LLMs' ability to understand and apply grammar knowledge in authentic teaching scenarios. CPG-EVAL both functions as a standardized assessment tool for evaluating generative models' instructional capabilities and offers a theoretical foundation for the future development of intelligent language education systems oriented toward language education.

\section{Related work}\label{sec:RelatedWork}

The development of benchmarks for measuring the evaluation of LLM capabilities has been actively pursued. Representative examples include GLUE \citep{wang2018GLUEMultitaskBenchmark} and SuperGLUE\citep{wang2019SuperGLUEStickierBenchmark} for assessing English language understanding, as well as CLUE \citep{xu2020CLUEChineseLanguage} and SuperCLUE \citep{xu2023superclue} for Chinese. Moving beyond natural language understanding, benchmarks such as MMLU\citep{hendrycks2020MeasuringMassiveMultitask} and BIG-bench \citep{srivastava2022beyond} have been developed to evaluate general knowledge and reasoning abilities. domain-specific benchmarks have emerged across various fields, including PubMedQA \citep{jin2019pubmedqa} in medicine, FINQA \citep{chen2021finqa}in finance, and LegalBench\citep{guha2023legalbench} in law. In the field of education, existing benchmarks include E-EVAL \citep{hou2024eval}, which targets K-12 educational knowledge in China.

However, none of these prior studies specifically address foreign language education. Although some benchmarks include items related to language, culture, or linguistics, these represent only a very small proportion of the total. For instance, C-EVAL \citep{huang2023CEVALMultilevelMultidiscipline}, one of the most prominent benchmarks for evaluating LLMs’ knowledge of Chinese, includes only a few items directly related to instructional content in Chinese learning as a second language. Therefore, it is difficult to argue that C-EVAL adequately reflects LLMs’ capabilities in the context of Chinese language teaching and learning.

Overall, existing benchmarks provide comprehensive evaluation frameworks for general language understanding, reasoning, and multilingual processing. However, no dedicated benchmark currently exists to assess the pedagogical grammar capabilities of LLMs. At a macro level, the absence of such evaluation tools may lead to policymakers misjudging the current capabilities of LLMs in language education, increasing the risk of misguided decisions. At a micro level, the lack of proper evaluation tools also exposes educators and learners to the risk of selecting inappropriate LLMs as teaching tools. Furthermore, LLM researchers and developers may not receive the necessary feedback to refine models for this domain, which could hinder the development of LLMs optimized for language education.

To fill this gap in language education evaluation benchmarks, we introduce a benchmark specifically designed to assess the pedagogical grammar knowledge of LLMs, enabling a more precise evaluation of their capabilities in foreign language instruction.

\section{The Construction of CPG-EVAL}\label{sec:TheConstructionofCPG-EVAL}
\begin{figure}[htbp]
\centering
\caption{Overall architecture of the CPG-EVAL benchmark.}
\includegraphics[width=\linewidth]{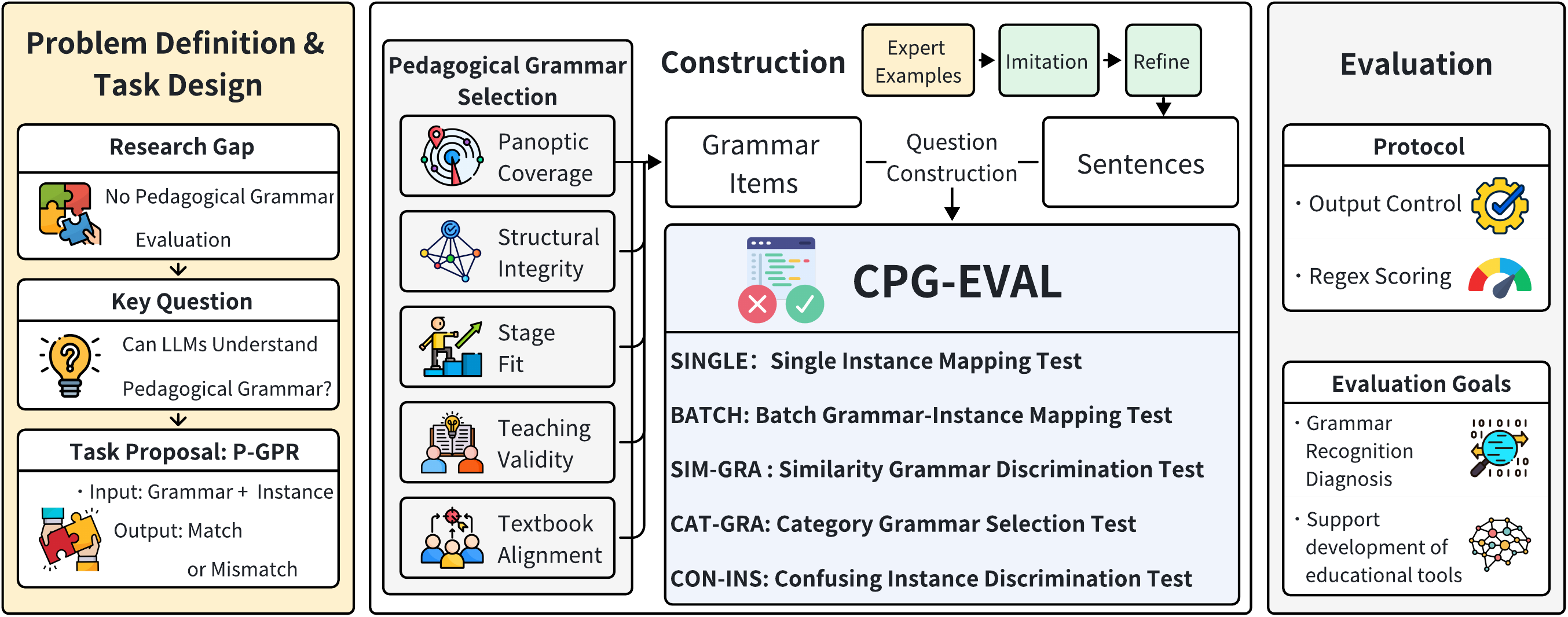}
\captionsetup{justification=raggedright,singlelinecheck=false}
\vspace{0.5em}
\begin{tablenotes}
\small
\item \textbf{Note.} For “Problem Definition \& Task Design,” see \autoref{sec:ProblemTask}; for “Grammar Items and Sentences,” see \autoref{sec:DataCollection}; for “Question Construction,” see \autoref{sec:QuestionConstruction}. For details on the evaluation procedure, refer to \autoref{sec:eval} and \autoref{sec:AnalysisDiscussion}.
\end{tablenotes}
  \label{fig:overall}
\end{figure}

This section describes in detail the conceptual framework and construction of the CPG-EVAL. \autoref{fig:overall} provides an overview.

\subsection{Problem Definition \& Task Design}\label{sec:ProblemTask}

The language-processing capabilities of LLMs are based on statistical learning methods, which differ substantially from human grammatical competence \citep{marcus2003AlgebraicMindIntegrating,tomasello2005ConstructingLanguageUsageBased,vaswani2017AttentionAllYou,brown2020language}. This difference implies that even though texts generated by LLMs may appear grammatically correct and logically coherent, the output primarily reflects the model's proficiency in predicting the next word rather than an understanding of pedagogical grammar principles. For instance, when completing the sentence "You \underline{\hspace{1cm}} my best friend," an LLM might correctly choose "are" not because it understands the pedagogical grammar rule of subject-verb agreement, but because this expression frequently appeared in its training data. Currently, there is no established method for measuring LLMs' understanding of pedagogical grammar. Consequently, a critical research gap remains unfilled, one which significantly influences the beneficial integration of LLMs in foreign language education: Can LLMs understand pedagogical grammar?

"Pedagogical grammar" refers to a meta-linguistic system designed to facilitate target language acquisition among second language learners \citep{odlin1994PerspectivesPedagogicalGrammar}. For example, phrases such as "nice and friendly," "big and comfortable," or "Turkish and English" can be abstracted into a pedagogical grammar rule described as "FORM: COMBINING TWO ADJECTIVES WITH ‘AND’"\citep{okeeffe2022EnglishGrammarProfile}. This abstraction exemplifies how pedagogical grammar generalizes rules underlying specific language instances.

From this perspective, any agent engaged in language teaching tasks—whether a human teacher or an artificial intelligence system—must recognize the correspondence between grammar items and the linguistic instances they describe. Specifically, the agent should accurately determine whether particular language instances (e.g., "nice and friendly") exemplify certain grammar items (e.g., FORM: COMBINING TWO ADJECTIVES WITH ‘AND’). Additionally, it must evaluate the appropriateness of language instances as examples of specific grammar items, distinguishing acceptable examples (FORM: COMBINING TWO ADJECTIVES WITH ‘AND’: "big and comfortable") from unacceptable ones (FORM: COMBINING TWO ADJECTIVES WITH ‘AND’: "you and me"). \cite{wang2025EvaluationLargeLanguage} terms this evaluative task "P-GPR" (Pedagogical Grammar Pattern Recognition) and formalizes it as follows:

\[
f\left(G_i, L_i\right) \rightarrow \{0, 1\}
\]

Here, Gi represents a {Grammar Item}, and Li denotes a  {Language Instance}(curly braces {} are used to indicate placeholder variables). The task requires determining whether Gi and Li correspond / match (→1) or do not correspond / mismatch (→0).
However, the basic P-GPR proposed by \cite{wang2025EvaluationLargeLanguage} evidently does not adequately capture the multidimensional understanding and application capabilities of LLMs regarding pedagogical grammar. Therefore, building upon the initial P-GPR modeling, this study proposes five types of tasks, which vary in difficulty and focus on different aspects of grammatical knowledge. Further details are provided in Section 3.2.3.
\subsection{Data Collection}\label{sec:DataCollection}
\subsubsection{\{Grammar Item\}}\label{sec:GrammarItem}
To transform basic task modeling into an operational evaluation benchmark, it is necessary to address issues related to the selection of the benchmark knowledge framework (Pedagogical Grammar Selection) and the construction of a corresponding example sentence database for grammar items.

The choice of pedagogical grammar is particularly critical, as it reflects the perspectives of a specific expert community and embodies the core values of the evaluation instrument. In the field of English as a Second or Foreign Language (ESL), The English Grammar Profile (O’Keeffe and Mark 2022) compiles over 1,200 statements of grammatical competence, each outlining what learners are expected to achieve at various CEFR levels. In the context of Teaching Chinese as a Second Language, the Chinese Grammar Learning Manual (Ying et al. 2022, hereafter abbreviated as CGLM) was developed by Chinese Testing International following the official publication of the Chinese Proficiency Grading Standards for International Chinese Language Education (National Language Commission 2021), commonly referred to as the "Standards." The CGLM is a systematically designed grammar syllabus textbook intended to support the implementation of the Standards and cater to the needs of global Chinese language learners.

Beyond the CGLM, numerous pedagogical grammar frameworks are available in various textbooks and supplementary materials. However, for our initial attempt at establishing a benchmark for teaching Chinese grammar knowledge specifically targeted at LLMs, we selected the CGLM as the knowledge core for the Chinese Pedagogical Grammar Evaluation (CPG-EVAL), due to the following advantages:

\begin{itemize}
    \item \textbf{Comprehensive Coverage}: The CGLM offers a grammar outline highly aligned with the Standards, covering all proficiency levels from beginner to advanced, and aligned with international examination systems such as HSK and YCT.
    \item \textbf{Architectural Integrity}: Grammar items in the CGLM are not listed in isolation; they are instead systematically categorized and integrated according to dimensions such as phonetics, vocabulary, semantics, and pragmatics, establishing explicit interconnections and a pedagogically sequenced progression of grammatical functions.
    \item \textbf{Stage Appropriateness}: The CGLM adjusts the cognitive load for learners at different proficiency levels, ensuring that grammar items at each level are both challenging and achievable.
    \item \textbf{Teaching Effectiveness}: Developed from over two decades of grammar teaching and assessment practices, the CGLM incorporates extensive empirical teaching data and classroom feedback, verifying the strong correlation between grammar points and teaching activities, and demonstrating its effectiveness.
    \item \textbf{Educational Compatibility}: The CGLM aligns closely with the grammar distribution patterns in mainstream international Chinese textbooks, such as "Developing Chinese," "New Practical Chinese Reader," and "HSK Standard Course," ensuring high compatibility between the grammar benchmark and textbook content.
\end{itemize}

In this study, we constructed a structured database based on the grammar items in the GLM. After excluding four items serving purely indexing purposes and without practical teaching content, we retained 739 grammar items of substantial pedagogical significance to form the basis of the CPG-EVAL grammar items.

\subsubsection{\{Language Instance\}}\label{sec:LangugeInstance} 
In the CGLM , each grammar item is accompanied by nine example sentences designed to illustrate the linguistic phenomena described by that grammar item. These example sentences have been thoroughly reviewed by the editorial committee of the CGLM and revised by Chinese linguists. To balance professionalism and open accessibility, this study employs synthetic data "equivalent to expert data in terms of its relationship to grammar items."

The synthesis process involves the following steps: 1) Using the official example sentences from the CGLM as templates, rewritten versions were generated by DeepSeek-v3; 2) Three native Chinese speakers, including the author, reviewed these sentences to ensure their compliance with Chinese linguistic conventions and to confirm that the synthetic data accurately preserved the relationship between the original expert examples and their respective grammar items; 3) The synthesized sentences were categorized in a database according to their corresponding grammar items. Through this method, a total of 739 grammar items × 9 sentences = 6651 synthetic sentences were obtained as language instances for constructing benchmark questions.

Additionally, we constructed highly confusing instances designed to be particularly challenging, aiming to further evaluate the LLMs' resistance to interference (see \autoref{sec:QuestionConstruction}).

\subsection{Question Construction}\label{sec:QuestionConstruction}

This section details the specific types of questions included in CPG-EVAL. Building upon Wang (2025)’s Single Instance Mapping Test, the present study expands the assessment methods to evaluate the multidimensional capabilities of LLMs.All actual prompts were written in Chinese; however, English is used in the following examples for ease of presentation.

\begin{enumerate}
    \item\textbf{SINGLE: Single Instance Mapping Test} SINGLE specifically assesses whether LLMs can accurately identify the correspondence between a single {Language Instance} and the indicated Grammar Item, following the basic format of P-GPR defined by \cite{wang2025EvaluationLargeLanguage}. SINGLE includes two subtasks: SINGLE-T (answer is T) and SINGLE-F (answer is F) .
        
        \textbf{Prompt template}: Does \{Language Instance\} contain \{Grammar Item\}?
        
        \textbf{Example}: Does the sentence “我不会弹钢琴” (I cannot play the piano) contain the grammar point [情态动词：会](Modal verb: huì——can, will)? If yes, output: T; if no, output: F.
        
        Answer: T
    
    \item\textbf{BATCH: Batch Grammar-Instance Mapping Test} BATCH assesses the ability of LLMs to determine the correspondence between multiple Language instances and the indicated Grammar Item. Compared to SINGLE, BATCH poses greater challenges, evaluating "complex context processing ability" (due to multiple sentences being involved) and "instruction-following ability," as it requires LLMs to individually determine the status of each sentence. BATCH also includes two subtasks: BATCH-T and BATCH-F.
        
        \textbf{Prompt template}: Does the following set of sentences contain the grammar point \{Grammar Item\}?
        
        \{Language Instance\} x9
        
        \textbf{Example}: Please evaluate sentences 1 through 9 one by one. If a sentence contains the grammar point [能愿动词：能] (Modal verb: néng – can/be able to), respond with T. If not, respond with F. Use only T or F for each sentence. No explanation is needed.
            \begin{enumerate}[label=\arabic*.]
                \item 天快黑了，我得回家了。 (It’s getting dark, I have to go home.)
                \item 我还得再想想。 (I still have to think it over.)    
                \item 这双鞋得一千多吧？ (These shoes must cost over a thousand, right?)   
                \item 你以后有想法得跟大家分享。 (In the future, if you have ideas, you must share them with everyone.)
                \item 开车的时候得注意安全。 (You have to be careful when driving.)
                \item 这个任务得你来完成，其他人不行。 (You must be the one to complete this task—no one else can.)
                \item 最近太忙了，今天我得好好休息。 (I’ve been too busy lately; today I  have to get a proper rest.)
                \item 这个项目得两周才能结束。 (This project will take two weeks to finish.)
                \item 你别担心，她一会儿准得回来。 (Don’t worry—she’s sure to come back in a while.)
            \end{enumerate}

    \item\textbf{SIM-GRA: Similarity Grammar Discrimination Test} SIM-GRA evaluates whether LLMs can distinguish between grammatically similar Grammar Items. The four distractor Grammar Items are derived by calculating semantic vector similarity using embedding models . This test challenges the ability of LLMs to perform fine-grained semantic distinctions despite superficial semantic similarities.
        
        \textbf{Prompt template}: Which of the following grammar points best matches the given  {Language Instance}?
        
        \textbf{Example}: Which grammar point does the sentence “他没有哥哥。” (He doesn't have an older brother.) best exemplify? Please answer using only the option letter.
            \begin{enumerate}[label=\Alph*.]
                \item “有”字句：表示附着：主语+动词+有+宾语("You"-construction: indicating attachment – Subject + Verb + 有 + Object)
                \item “有”字句：表示领有("You"-construction: indicating possession)
                \item “有”字句：表示存在、具有：主语+有+着+宾语("You"-construction: indicating existence/possession with 着)
                \item“有”字句：表示比较("You"-construction: indicating comparison)
                \item“有”字句：表示存在("You"-construction: indicating existence)
            \end{enumerate}
        
        Answer: B
        
    \item\textbf{CAT-GRA: Category Grammar Selection Test} CAT-GRA specifically evaluates the ability of LLMs to differentiate among Grammar Items within the same grammatical category. In CAT-GRA, all Grammar Items belonging to the same grammatical category as the correct answer are provided as options. This approach examines the ability of LLMs to make fine-grained distinctions between grammatical items within the domain of pedagogical grammar. Some questions contain up to 56 options, significantly testing the model's capacity for complex and extended context processing.
    
        \textbf{Prompt template}: Which of the following grammar points best matches the  {Language Instance}?
        
        \textbf{Example}：Which grammar point does the sentence “苹果是红的。” (The apple is red.) best exemplify?
            \begin{enumerate}[label={}, leftmargin=*]
                \item1063.“是”字句：表示等同或类属("Shi"-construction: expressing equivalence or class membership)
                
                \item1064.“是”字句：表示说明或特征("Shi"-construction: indicating description or characteristic)
                
                \item1065.“是”字句：表示存在("Shi"-construction: expressing existence)
                
                \item3113.用“是”强调(Using “shi” for emphasis)
            \end{enumerate}
            Answer: 1064
    \item\ \textbf{CON-INS: Confusing Instance Discrimination Test} 
    CON-INS evaluates whether LLMs can correctly discriminate when a Language Instance contains linguistic forms similar to a Grammar Item (referred to as a Confusing Instance), but which do not actually correspond to that grammar item.CON-INS includes two subtasks: CON-INS-F10 (all instances do not contain the targeted grammar point) and CON-INS-T5F5 (half of the instances contain the targeted grammar point, half do not). CON-INS can be viewed as a variant of BATCH, increasing the difficulty at the Language Instance level by requiring LLMs to resist interference from linguistic similarities.The following example is from the CON-INS-F10 sub-set, where none of the instances align with the targeted grammar item.
    
        \textbf{Prompt template}: Does the following set of sentence contain the grammar point {Grammar Item}?
        
        \textbf{Example}: Please evaluate sentences 1 through 10 one by one. If a sentence contains the grammar point [Fixed structure: 在……看来] (in someone’s opinion, from the someone’s perspective), respond with T. If not, respond with F.
        \begin{enumerate}[label=\arabic*.] 
            \item 他在办公室里工作。(He works in the office.)
            \item 看来今天的天气不错。(It seems the weather is nice today.)
            \item 在医生看来，定期体检有助于预防疾病。(In the doctor’s opinion, regular check-ups help prevent illness.)
            \item 在我看来，健康的饮食比运动更重要。(In my opinion, a healthy diet is more important than exercise.)
            \item 他在图书馆里看书。(He is reading in the library.)
            \item 在老板看来，团队合作是成功的关键。(In the boss’s view, teamwork is key to success.)
            \item 在我看来，这部电影的结局有些出人意料。(In my opinion, the ending of this movie is a bit unexpected.)
            \item 她在厨房里做饭。(She is cooking in the kitchen.)
            \item 在长辈看来，传统节日是家庭团聚的好时机。(From the elders’ perspective, traditional holidays are a good time for family reunions.)
            \item 她在教室里学习。(She is studying in the classroom.)
        \end{enumerate}

\textbf{Note.}Confusing Instances primarily involve two types: 
    \begin{enumerate}
        \item Instances containing words identical in pronunciation or form to a Grammar Item but differing in grammatical meaning. For example: Modal verb: 会 (can, will) vs. 会议 (meeting); Degree adverb: 可 (can) vs. 可持续 (sustainable). 
        \item Instances partially sharing form or pronunciation but differing in grammatical meaning. For example: Fixed phrase A着A着(fixed pattern indicating a continuing action leading to a change of state, e.g., “说着说着他就哭了” – As he was speaking, he suddenly started crying) vs. 正忙着(progressive aspect marker 着 used in a standard verb phrase, e.g., She is currently busy); Colloquial structure 什么X的Y的 (colloquial construction used to vaguely list or generalize different categories of things, implying no clear distinction, e.g., “什么你的我的，都是大家的” (Yours, mine — it’s all shared by everyone.) vs. 什么adj的noun (A generalizing “什么” happens to precede an “adjective + noun” structure that contains the particle “的.”), e.g.她正在学习什么新的语言 (She is learning some new language)
    \end{enumerate}
    Initially, 314 grammar items were selected following these two criteria, and DeepSeek-v3 was employed to generate sentences featuring similar linguistic forms but not aligned with the targeted Grammar Items. Two native-speaking Chinese language teachers specializing in teaching Chinese as a second language then screened and revised the sentences.
\end{enumerate}

For the total number of questions in each question bank, see the "Number of Questions" column in \autoref{tab:questions}.

\begin{table}[htbp]
\centering
\begin{minipage}[t]{0.48\linewidth}
\centering
\begin{threeparttable}
\caption{Distribution of Question Items by Type in the CPG-EVAL Benchmark.}
\label{tab:questions}
\begin{tabular}{lr}
\toprule
\textbf{Question Bank} & \textbf{Number of Questions} \\
\midrule
SINGLE-T        & 6,651  \\
SINGLE-F        & 11,968 \\
BATCH-T         & 739    \\
BATCH-F         & 11,968 \\
SIM-GRA         & 6,651  \\
CAT-GRA         & 5,283  \\
CON-INS-F10     & 314    \\
CON-INS-T5F5    & 314    \\
\bottomrule
\end{tabular}
\vspace{0.5em}
\begin{tablenotes}
\small
\item \textbf{Note.} SINGLE and BATCH test grammar recognition in single or multiple sentences, with subtypes T/F indicating presence or absence of the grammar item. SIM-GRA and CAT-GRA assess fine-grained grammar selection among similar or same-category items. CON-INS tasks evaluate robustness against confusing instances. See \autoref{sec:QuestionConstruction} for detailed definitions and examples.
\end{tablenotes}
\end{threeparttable}
\end{minipage}
\hfill
\begin{minipage}[t]{0.48\linewidth}
\centering
\begin{threeparttable}
\caption{Model specifications: parameters and access method.}
\label{tab:model-specs}
\begin{tabular}{lcc}
\toprule
\textbf{Model} & \textbf{\textit{\#Parameters}} & \textbf{Access} \\
\midrule
Qwen2.5-7B-Instruct & 7B & Weights \\
internlm2\_5-7b-chat & 7B & Weights \\
Llama-3.1-8B-Instruct & 8B & Weights \\
GLM-4-9B-Chat & 9B & Weights \\
Qwen2.5-70B-Instruct & 70B & Weights \\
DeepSeek-V3-250324 & 660B MoE & API \\
GPT-4o-2024-08-06 & Undisclosed & API \\
GPT-4o-mini-2024-07-18 & Undisclosed & API \\
Doubao-1-5-pro-32k-250115 & Undisclosed & API \\
Doubao-1-5-lite-32k-250115 & Undisclosed & API \\
Qwen2.5-MAX-250409 & Undisclosed & API \\
\bottomrule
\end{tabular}
\vspace{0.5em}
\begin{tablenotes}
\small
\item \textbf{Note.} For more information about models, see the following references: Qwen \citep{qwen25}, GLM \citep{teamglm2024ChatGLMFamilyLarge}, internlm2\_5\citep{cai2024InternLM2TechnicalReport}, Llama \citep{grattafiori2024Llama3Herd}, DeepSeek\citep{deepseek-ai2025DeepSeekV3TechnicalReport}, GPT \citep{openai2024Gpt4o20240806,openai2024Gpt4oMini20240718}, Doubao\citep{doubaoteam2025Doubao15Pro}.
\end{tablenotes}
\end{threeparttable}
\end{minipage}
\end{table}

\section{Evaluation}\label{sec:eval}
\subsection{Setup and Models}

We employed CPG-EVAL to assess multiple open-source and proprietary LLMs. The evaluation was conducted under a zero-shot setting, meaning that no additional examples or supplementary information beyond the questions and choices introduced in \ref{sec:QuestionConstruction} were provided. In this way, we aimed to test the intrinsic knowledge and reasoning capabilities of the models. The list of evaluated models is provided in \autoref{tab:model-specs}.

To calculate scores objectively, we utilized regular expressions to extract the answers from the LLM outputs. To reduce noise that could affect statistical accuracy, explicit instructions such as “Please respond using only the letter corresponding to the correct answer.” were embedded in the prompts to enforce output control, thereby ensuring consistency and clarity in the LLM-generated responses. Finally, we extracted answers from the model responses using regex scoring and calculated each model’s scores across the five evaluation categories. 

\subsection{Results}

The performance of each model across the five evaluation categories is summarized in \autoref{tab:model-eval-results}.In CPG-EVAL, each BATCH question comprises 9 sub-questions, while each CON-INS question comprises 10 sub-questions. Accuracy is calculated independently at the sub-question level. For example, if a CON-INS question has the correct answer “TTTFFTTFFF” and the model responds with “TTTFFTFTTT”, the model correctly answers 6 out of 10 sub-questions, thus scoring 0.6 points (with a full score of 1 point per question).

\begin{table}[htbp]
\centering
\begin{threeparttable}
\caption{Evaluation results across five test types (SINGLE, BATCH, SIM-GRA, CAT-GRA, CON-INS) and average scores.}
\label{tab:model-eval-results}
{\begin{tabular}{lccccccccc}
\toprule
 & \multicolumn{2}{c}{SINGLE} & \multicolumn{2}{c}{BATCH} & \multirow{2}{*}{SIM-GRA} & \multirow{2}{*}{CAT-GRA} & \multicolumn{2}{c}{CON-INS} & \multirow{2}{*}{Average} \\
\cmidrule(r){2-3} \cmidrule(r){4-5} \cmidrule(r){8-9}
\multicolumn{1}{c}{\textbf{Model}} & T & F & T*9 & F*9 & & & F*10 & T*5\&F*5 & \\
\midrule
RANDOM & .500 & .500 & .500 & .500 & .200 & .100 & .500 & .500 & 0.413 \\
Doubao-1.5-pro & .982 & .939 & \textbf{.985} & \textbf{.943} & \textbf{.940} & \textbf{.928} & \textbf{.781} & .919 & \textbf{.927} \\
GPT-4o & .972 & .930 & .926 & .933 & .919 & .904 & .780 & .912 & .910 \\
DeepSeek-v3 & .943 & \textbf{.952} & .944 & .902 & .917 & .875 & .757 & .904 & .899 \\
Qwen2.5-Max & \textbf{.984} & .897 & .854 & .917 & .905 & .867 & .775 & \textbf{.920} & .890 \\
Doubao-1.5-lite & .950 & .926 & .922 & .893 & .900 & .851 & .684 & .878 & .876 \\
Qwen2.5-72B & .943 & .943 & .942 & .902 & .894 & .847 & .665 & .852 & .873 \\
GPT-4o-mini & .915 & .912 & .956 & .727 & .863 & .798 & .537 & .824 & .817 \\
Qwen2.5-7B & .897 & .926 & .720 & .780 & .845 & .619 & .582 & .717 & .761 \\
GLM-4-9B & .951 & .772 & .930 & .535 & .775 & .710 & .368 & .723 & .720 \\
LLaMA-3.1-8B & .977 & .751 & .810 & .543 & .812 & .611 & .392 & .728 & .703 \\
InternLM2.5-7B & .970 & .753 & .742 & .316 & .923 & .542 & .359 & .624 & .654 \\
Ave.Question & .951 & .873 & .882 & .730 & .877 & .760 & .582 & .803 & .807 \\
\bottomrule
\end{tabular}}
\vspace{0.5em}
\begin{tablenotes}
\small
\item \textbf{Note.} SINGLE = Single Instance Mapping Test; BATCH = Batch Grammar-Instance Mapping Test; SIM-GRA = Similarity Grammar Discrimination Test; CAT-GRA = Category Grammar Selection Test; CON-INS = Confusing Instance Discrimination Test. For detailed task descriptions, see \autoref{sec:QuestionConstruction}. For model specifications, refer to \autoref{tab:model-specs}.
\end{tablenotes}
\end{threeparttable}
\end{table}

\section{Analysis and Discussion}\label{sec:AnalysisDiscussion}
\subsection{Results of SINGLE and BATCH}

In the CPG-EVAL evaluation framework, the SINGLE task represents the simplest scenario by merely requiring LLMs to judge whether instructional grammatical descriptions match particular language instances. Therefore, it offers an intuitive reflection of an LLM's capability in recognizing the linguistic phenomena defined by pedagogical grammar rules.
Firstly, the overall trend from the SINGLE task demonstrates that a majority of tested LLMs have strong capabilities in correctly identifying positive instances (SINGLE-T, i.e., grammar match), yielding an average accuracy of 95.1\%. Models such as Qwen2.5-Max (98.4\%), Doubao-1.5-pro (98.2\%), and LLaMA-3.1-8B (97.7\%) show the highest performance. Notably, even LLaMA-3.1-8B and InternLM2.5-7B, achieve a high-performance rate (97.0\%), surpassing several larger-scale counterparts.

However, performance significantly declines for most LLMs when handling negative instances (SINGLE-F, i.e., no grammar match), with an average reduction of 7.8\%. This decline is particularly pronounced in the cases of GLM-4-9B, LLaMA-3.1-8B, and InternLM2.5-7B, whose accuracies drop by 17.9\%, 22.6\%, and 21.7\%, respectively. Such performance patterns clearly indicate that numerous models conspicuously struggle in tasks involving grammar mismatch identification, exhibiting false-positive errors. One notable exception is DeepSeek-v3, which achieves higher accuracy for negative instances (SINGLE-F 95.2\%) than positive ones (SINGLE-T 94.3\%), suggesting an intentionally cautious judgment strategy that tends toward conservatively deciding "absence of a specific grammatical phenomenon." This approach represents a rare success among the models tested for negative grammar identification.

When the task becomes more complex—transitioning from SINGLE instances to BATCH tasks that require the simultaneous processing of multiple linguistic instances—differences among the tested LLMs become even more pronounced. Most models maintain relatively high performance when identifying positive batch examples (BATCH-T), averaging 88.2\%, but display a marked and substantial decline in average accuracy (73.0\%) for negative batch tasks (BATCH-F), equating to a significant average reduction of 15.2\%. InternLM2.5-7B demonstrates an especially large decrease, falling from 74.2\% on BATCH-T to 31.6\% on BATCH-F, significantly below the baseline of 50\%. Similar declines are evident in GLM-4-9B (from 93\% to 53.5\%) and LLaMA-3.1-8B (from 81\% to 54.3\%). Additionally, GPT-4o-mini, which ranks near the top on positive batch tasks (95.6\% on BATCH-T), decreases to 72.7\% for negative batch tasks, further highlighting the significant limitations some models exhibit in negative batch instance identification.

By contrast, larger-scale models such as Doubao-1.5-pro (98.5\% on BATCH-T, 94.3\% on BATCH-F), GPT-4o (92.6\%, 93.3\%), and DeepSeek-v3 (94.4\%, 90.2\%) demonstrate consistently robust judgment capabilities and maintain an optimal balance, clearly underscoring the advantage large-scale models have in terms of performance stability and generalization capabilities across increasingly complex task types.

Overall, analysis of SINGLE and BATCH tasks reveals four critical observations:

\begin{enumerate}[label=(\alph*)]
    \item In simple (SINGLE) tasks, some smaller-scale LLMs can achieve performance approaching or even surpassing those of certain larger-scale models.
    \item There exist noticeable strategic differences among LLMs in determining the presence or absence of grammatical items. Some models adopt a more aggressive response strategy that readily yields affirmative responses, while others are more conservative.
    \item However, as tasks grow more complex (BATCH tasks), large-scale models clearly outperform smaller-scale models in recognizing negative instances, demonstrating substantially enhanced stability and accuracy. Smaller-scale models often fall below the random baseline.
    \item A pervasive difficulty among LLMs lies specifically in identifying negative grammatical instances (i.e., instances that do not match grammar rules), with certain models displaying significant false-positive issues that deserve close attention.
\end{enumerate}

These results offer the following implications:\textbf{1)} Current LLMs applied to real-world pedagogical contexts are likely to misclassify sentences as containing specific grammatical phenomena which are not present. \textbf{2)} For practical use as grammar teaching tools, LLM selection strategies must adjust to different task complexities and requirements. 

This reinforces the importance of a systematically multi-tiered evaluation framework for diagnosing and better understanding gaps and limitations in LLMs' capabilities.

\subsection{From the Results of SIM-GRA and CAT-GRA}

If LLMs are to be used in foreign language teaching, their fine-grained grammatical discrimination ability is essential. The two types of tasks explored in this study, SIM-GRA and CAT-GRA, share considerable similarity: both require LLMs to identify the grammatical rule that best describes a given linguistic instance. The primary distinction between these two task types lies in how distractors are generated, either semantically or grammatically. 

Overall, Models achieve an average accuracy of 87.7\% on tasks with semantic-similarity-based distractors (SIM-GRA), notably outperforming their 76.0\% accuracy on tasks with structured grammatical-category-based distractors (CAT-GRA). larger-scale models such as GPT-4o and DeepSeek-V3 tend to demonstrate stronger resistance to interference stemming from both ""literal similarities"" and ""categorical similarities"" when compared to smaller-scale models. One intriguing example is the InternLM2.5-7B model, which achieves excellent performance in SIM-GRA tasks (92.3\%), surpassing some larger-scale models, yet experiences a considerable performance drop in CAT-GRA tasks (54.2\%). This underscores the point that literal and categorical semantic similarities challenge LLMs in fundamentally different ways. In other words, the more specialized and professionally structured the language teaching task, the greater the challenge for these models.

The largest language models consistently achieve high performance across both task types, reflecting clear advantages in both robustness and stability. For instance, Doubao-1.5-pro demonstrates leading performance in both SIM-GRA (94.0\%) and CAT-GRA (92.8\%), while GPT-4o (91.9\% in SIM-GRA, 90.4\% in CAT-GRA) and DeepSeek-v3 (91.7\% SIM-GRA, 87.5\% CAT-GRA) also show high performance. These results affirm that models with substantial scale and comprehensive training achieve higher degrees of robustness and generalization when faced with fine-grained grammatical distinctions and specialized interference.

In contrast, smaller-scale models such as InternLM2.5-7B and LLaMA-3.1-8B significantly underperform in CAT-GRA tasks—achieving scores of only 54.2\% and 61.1\%, respectively—which are considerably lower than their SIM-GRA accuracy rates (InternLM2.5-7B: 92.3\%; LLaMA-3.1-8B: 81.2\%).

In conclusion, the current generation of LLMs commonly encounters difficulties in handling detailed, specialized interference within the structured knowledge systems of pedagogical grammar. This limitation is especially pronounced for smaller-scale models, indicating that for fine-grained grammar discrimination tasks within foreign language teaching scenarios, selecting larger language models is currently advisable due to their consistently superior performance.

\subsection{From the Results of CON-INS}

When responding to CON-INS tasks, LLMs particularly need to focus on overcoming interference arising from linguistic form when making judgements at the grammatical-functional level.

In the CON-INS-F10 setting, the task specifically examines whether models can completely exclude interference from multiple confusing instances. Experimental results show that the top-performing Doubao-1.5-pro (78.1\%), followed closely by GPT-4o (78.0\%) and Qwen2.5-Max (77.5\%), exhibit highly similar performance levels. In contrast, smaller-scale models, such as GPT4o-mini (53.7\%) and Qwen2.5-7B (58.2\%), achieve accuracy rates only slightly above the level of random guessing. The accuracy for internlm2\_5-7b (35.9\%) and glm-4-9b (36.8\%) is substantially lower than the random baseline. These results reveal that smaller models are particularly susceptible to interference from highly confusing instances, potentially producing systematic biases.

Under the balanced CON-INS-T5F5 condition, where the number of confusing instances (sentences that should be answered with an 'F') was reduced from 10 (as in CON-INS-F10) to 5, the accuracy for all tested models improved considerably. Qwen2.5-Max achieved the best performance, with an accuracy of 92.0\%, followed by Doubao-1.5-pro (91.9\%), GPT-4o (91.2\%), and DeepSeek-V3 (90.4\%), which all exhibited high performance. Overall metrics clearly demonstrate that the reduction in the number of confusing instances significantly influences the models' robustness against interference, with all models demonstrating clear gains in accuracy as task difficulty decreases.

Moreover, each model performed significantly better in the T5F5 condition compared to its corresponding F10 condition (for example, internlm2\_5-7b improved from 35.9\% in F10 to 62.4\% in T5F5). This further suggests the strong negative impact of the quantity of confusing instances on the models’ ability to resist interference.

In conclusion, the CON-INS evaluations reaffirm the potential and advantages of large-scale LLMs in language-teaching assistance and detailed grammatical discrimination tasks. At the same time, these tests highlight the challenges faced by smaller-scale models in tasks involving highly confusing linguistic forms. This indicates the ongoing difficulties in effectively using these models as language-teaching assistants. The observed performance gap underscores the importance of efforts targeted at enhancing grammatical discrimination and interference resistance capabilities in small-scale, cost-effective language models for educational scenarios.

\section{Conclusion}

This study introduced the conceptual foundation, construction method, and evaluation of CPG-EVAL, the first benchmark specifically designed to assess the capabilities of LLMs in foreign language teaching scenarios. Using the grammatical knowledge framework provided by the 'Chinese Grammar Learning Manual' (CGLM), complemented by human-refined synthetic datasets, CPG-EVAL encompasses five distinct yet complementary task types: SINGLE, BATCH, SIM-GRA, CAT-GRA, and CON-INS. These tasks are designed to test the critical capabilities required of LLMs in grammar instruction contexts, including grammar recognition, linguistic instance differentiation, and appropriate structured output.

Several key findings emerged from extensive evaluations of multiple LLMs:
\begin{enumerate}[label=(\arabic*)]
    \item Task Complexity and Model Scale:   1) Simple tasks: In straightforward scenarios such as the SINGLE task, smaller-scale LLMs sometimes approach or even exceed the performance of larger-scale models. This indicates that smaller models could represent a cost-effective option for simpler teaching scenarios.2) Complex tasks: As task complexity increases—such as the transition to BATCH tasks—the performance of smaller models declines significantly, demonstrating that smaller-scale models typically do not meet practical application requirements for complex instructional tasks. Therefore, caution is advised when adopting smaller LLMs in scenarios requiring the processing of multiple sentences at a time.
    
    \item Differences in Judgment Strategies across Models:   LLMs exhibit clear differences regarding decision-making strategies. Specifically, some models display a tendency towards affirmative judgment (more aggressive), while others adopt a more conservative approach characterized by frequent negative judgments. In practical settings, such inherent differences may influence model reliability and applicability.
    
    \item Difficulties with Negative Instance Recognition:   It was frequently observed that LLMs struggle to correctly identify negative grammar instances—examples that do not include a specific grammar item. Some models frequently misclassify negative instances as positives. In real-world language instruction, this indicates that current-generation LLMs risk misclassifying sentences that do not exemplify teaching content as valid or relevant; thus, careful oversight by users remains essential.
    
    \item Relationship between Model Scale and Resistance to Disturbance: Larger-scale models clearly demonstrate superior robustness against semantic interference, specifically literal or categorical similarity. However, even the most advanced models face significant challenges in handling structured grammatical interference defined by specialized grammatical frameworks. This illustrates that more specialized language-teaching tasks place greater demands on model performance. Consequently, for sensitive, fine-grained grammar discrimination tasks without additional fine-tuning, larger-scale models should be preferred. 
    
    \item Form-related interference: In CON-INS tasks, large models such as Doubao-1.5-pro and GPT-4o stand out in effectively resisting disturbances related to linguistic form. In contrast, smaller models are highly susceptible to interference arising from closely similar instances, frequently causing significant deterioration in performance. Nevertheless, although larger-scale models demonstrate clear advantages, experimental results suggest there remains substantial room for improvement in terms of judgment accuracy.
    
    \item Importance of a Comprehensive Evaluation Framework:   A systematic and multi-layered evaluation framework is crucial for diagnosing and understanding the gaps and limitations in LLMs’ capabilities in realistic applications. Such frameworks facilitate the choice of LLMs, their optimization in instructional applications and improved outcomes in language teaching.
\end{enumerate}

We firmly believe that the sound and sustainable integration of LLMs in foreign language education cannot be dissociated from academic research. The original rationale for the current study—to bridge the existing gap between language education specialists and the capabilities of LLMs through the creation of CPG-EVAL—reflects our ambition to support such interdisciplinary effort. With this first initiative in the field, we hope that CPG-EVAL will modestly contribute towards systematically assessing the capacities of LLMs as instructional tools. The conceptual discussion of CPG-EVAL represents an initial step toward theorizing the evaluation of LLM capabilities from the perspective of foreign language education. Additionally, the empirical findings and analyses provided by this study constitute valuable empirical evidence for shaping ""next-generation foreign language education integrated with AI."" Accordingly, by conducting grammar recognition diagnostics, we aim to help educators and researchers develop an accurate understanding of current LLM capabilities and thereby avoid excessive expectations or anxiety about AI. Moreover, this work seeks to offer actionable insights for future efforts in aligning and fine-tuning instructional content, as well as in advancing research on educational tools grounded in LLM technologies.

Nonetheless, we clearly recognize that benchmark development is never static. Given the rapid advancement of artificial intelligence technologies along with their application domains, we emphasize the necessity to continuously refine benchmark designs from diverse angles, including target instructional language, knowledge structure, problem formulations, and specific educational scenarios. Further exploration in these directions is the topic of ongoing work.

\bibliographystyle{chicago}
\bibliography{references}  

\end{CJK*}
\end{document}